\documentclass[
]{ceurart}

\usepackage{listings}
\usepackage{tcolorbox}
\usepackage{algorithm}
\usepackage{algorithmic}
\usepackage[table,dvipsnames]{xcolor}
\definecolor{lightred}{RGB}{255,200,200}
\definecolor{lightgreen}{RGB}{200,255,200}
\definecolor{myblue}{rgb}{0.8,0.9,1}
\definecolor{lightgray}{gray}{0.9}

\newcommand{\greenup}{\textcolor{green}{\boldmath$\uparrow$}}
\newcommand{\reddown}{\textcolor{red}{\boldmath$\downarrow$}}
\begin{document}

\copyrightyear{2026}
\copyrightclause{Copyright for this paper by its authors.
  Use permitted under Creative Commons License Attribution 4.0
  International (CC BY 4.0).}

\conference{1st Streaming Continual Learning Bridge at AAAI26, January 21, 2026, Singapore.}

\title{SOLAR: A Self-Optimizing Open-Ended Autonomous Agent for Lifelong Learning and Continual Adaptation}


\author[1,2]{Nitin Vetcha}[%
orcid=0009-0003-6542-324X,
email=nitinvetcha@iisc.ac.in,
url=https://github.com/nitinvetcha/,
]
\cormark[1]
\address[1]{Department of Ophthalmology, Yong Loo Lin School of Medicine, National University of Singapore, Singapore}
\address[2]{Department of Computational and Data Sciences, Indian Institute of Science, Bangalore, Karnataka, India}

\author[1]{Dianbo Liu}[%
orcid=0000-0002-3042-9161,
email=dianbo@nus.edu.sg,
url=https://www.asintelligence.xyz/,
]


\cortext[1]{Corresponding author.}

\begin{abstract}
Despite the remarkable success of large language models
(LLMs), they still face bottlenecks while deploying in dy
namic, real-world settings with primary challenges being
concept drift and the high cost of gradient-based adapta
tion. Traditional fine-tuning (FT) struggles to adapt to non
stationary data streams without resulting in catastrophic for
getting or requiring extensive manual data curation. To ad
dress these limitations within the streaming and continual
learning paradigm, we propose the Self-Optimizing Lifelong
Autonomous Reasoner (SOLAR) which is an open-ended au
tonomous agent that leverages parameter-level meta-learning
to self-improve, treating model weights as an environment
for exploration. It initiates the process by consolidating a
strong prior over common-sense knowledge making it effective for transfer-learning. By utilizing a multi-level reinforcement learning approach, SOLAR autonomously discovers adaptation strategies, enabling efficient test-time adaptation to unseen domains. Crucially, SOLAR maintains an
evolving knowledge base of valid modification strategies,
implicitly acting as an episodic memory buffer to balance
plasticity (adaptation to new tasks) and stability (retention
of meta-knowledge). Experiments demonstrate that SOLAR
outperforms strong baselines on common-sense, mathemati
cal, medical, coding, social and logical reasoning tasks, marking a significant step toward autonomous agents capable of
lifelong adaptation in evolving environments.
\end{abstract}


\maketitle

\section{Introduction}
\label{sec:intro}

Large Language Models (LLMs) possess remarkable emer
gent abilities due to massive pretraining. However, deploy
ing them in streaming environments reveals a critical weak
ness which is the inability to adapt to non-stationary data
distributions (concept drift) without expensive retraining or
human intervention. While Parameter-Efficient Fine-Tuning
(PEFT) techniques like LoRA (Hu et al. 20222) reduce the
parameter update volume, they remain static solutions that
do not inherently address the stability-plasticity dilemma
central to Continual Learning (CL).
Existing adaptation strategies often rely on generic, hand
crafted heuristics that fail to generalize across the shifting
temporal dependencies of real-world streams. This disconnect necessitates a system that can not only adapt parameters on the flybut also learn how to adapt based on accumulating
experience. We propose that the high-dimensional weight
space of an LLMcontains rich meta-knowledge that, if navi
gated autonomously, can yield bespoke adaptation strategies
for novel tasks. This motivates our primary research question:

\begin{tcolorbox}[colback=red!5!white,colframe=red!5!white]
\textcolor{red}{\textbf{RQ}: Can LLMs learn to modify their internal representation space autonomously to handle concept drift, analogous to how humans assimilate and restructure knowledge in lifelong learning scenarios?}
\end{tcolorbox}

\noindent To answer this, we investigate the cognitive science of life
long learning. As humans, we do not merely memorize new
data, instead we restructure our internal schematics in order
to be able to accommodate new information while simultaneously retaining prior heuristics. This process is what has
essentially enabled humans to navigate non-stationary environments. For instance, a student adapts their study strategy based on the nature of a new subject (plasticity) without
unlearning how to study generally (stability). Current LLM
adaptation, by contrast, is often rigid since models consume
task data “as-is”, failing to develop bespoke internal trans
formation strategies.
To replicate this cognitive flexibility, we introduce SOLAR
(Self-Optimizing Lifelong Autonomous Reasoner). It functions as a meta-learning agent that decouples rapid task
adaptation (streaming machine learning) from long-term
strategy retention (continual learning). By discovering and
validating parameter-level modifications, SOLAR enables
efficient adaptation to unseen tasks while populating a persistent knowledge base to mitigate catastrophic forgetting.
This work thus bridges the gap between static parameter
generation and dynamic, lifelong self-evolution. Further
more, by grounding the search space in neural network
weights, we target generalized principles of model capability rather than task-specific memorization. Just as scaling
laws (Kaplan et al. 2020) predict performance based on size,
we posit that predictable weight-modification patterns exist that allow for rapid, data-efficient adaptation to concept
drift, minimizing the lag between detecting a distributional
shift and deploying an updated model.
The remainder of this paper is organized as follows. In Sec
tion 2, we highlight the motivation for our approach in detail.
Section 3 presents the literature survey conducted, Section 4
contains the methodology with implementation specifics in
Section 5. Experimental results are provided in Section 6
and in Section 7, we present our concluding remarks.

\section{Motivation}
\label{sec:motiv}
Our primary motivation stems from human psychology and pedagogy. For example, consider a human student who is preparing to take an end-sem examination of a machine learning course. Quite often, students tend to rely on their prior prepared notes for preparation. These notes are often derived from the lecture content, textbooks or information available on the internet. Thus instead of relying on the raw content, students assimilate and rewrite the information in the form of notes as per their own intrinsic reasoning skill and aptitude. This improves the capability of students to comprehend the content better and therefore respond well to the exam questions. This phenomenon of reinterpreting and augmenting external knowledge in a way that is easier to understand as well as developing the necessary skill-sets
is not limited to just taking exams, but seems to be universally true of human learning across tasks. Furthermore, depending on one's interests, humans assimilate information in different ways - some might condense the information into a visual diagram, some into text, or some might rely more on concrete mathematical descriptions. Such restructuring or development of internal knowledge as well as assimilation or rewriting of external information, as part of the learning process is in contrast with how LLMs undergo currently training and adaptation. Given a new task, current LLMs consume and learn from the task data "as-is" via finetuning or in-context learning. The issue with this, just like in the human setting, such data may not be in an optimal format (or volume) for learning, or there might not be the relevant skill-set developed to learn it and current approaches do not enable models to develop bespoke strategies for how to best transform themselves internally or even learn from their training data. In this work, we therefore investigate the question as to if it is possible for even LLMs, analogous to humans, to suggest strategies by themselves which can enable them to perform better on a given task. 

\vspace{0.5cm}
\noindent 
A secondary source of motivation as to why we ground our strategy search space in the neural network weights is because unlike task-specific knowledge, the weight-level meta-knowledge represents generalized principles about how neural network parameters relate to model capabilities, thereby providing crucial insights for self-evolving agents. There are several prior research works which have shown that there exists a positive correlation between types of neural network weight patterns and downstream model performance characteristics. For example, scaling laws research \cite{kaplan2020scaling} has demonstrated that there are predictable relationships between model size and performance. Similarly, structured sparsity learning gives an indication so as to how particular weight patterns can be useful for developing more efficient representations \cite{wen2016learning}.

\section{Related Work}
\label{sec:lit_survey}

\textbf{Test-Time Training} (TTT) is a recently emerging class of approaches which updates model weights at inference time using techniques such as input perplexity or cross-entropy minimization on only unlabeled test data enabling self supervised enhancement of LLM performance \cite{hu2025testtimelearninglargelanguage, hu2025slotsamplespecificlanguagemodel} or via reinforcement learning by utilizing the priors in the pre-trained models \cite{zuo2025ttrltesttimereinforcementlearning} or by using reflection and verifier-driven sample selection \cite{moradi2025continuousselfimprovementlargelanguage, lee2025reviselearningrefinetesttime} or by using a task-specific curriculum \cite{hübotter2025learningjobtesttimecurricula} or by using a mixture-of-expert based model merging \cite{bertolissi2025localmixturesexpertsessentially}. An alternative approach is to scale inference compute at test time as well using for example ensemble approaches such as majority voting. While test-time approaches is a promising option, such a computational overhead might not be necessary always and it often fails in cases where data is scarce or quality of unlabeled data is poor. 

\vspace{0.5cm}
\noindent \textbf{Adversarial Fine Tuning} is another emerging class of techniques where in two LLM instances are made to debate with each other about a topic or one instance serves as a challenger or teacher and the other instance serves as a solver or student to generate synthetic data, either from unlabeled prompts or even from scratch itself and use approaches like majority voting to create pseudo-labels which can further be used for updating model's knowledge accordingly \cite{yang2024syntheticcontinuedpretraining, wang2025selfupdatablelargelanguagemodels, wang2025lokilowdamageknowledgeimplanting}. This can also be done by some additional fine-tuning using information which is available in the LLM's context as well \cite{park2025textitnewnewssystem2finetuning} similar to knowledge distillation. Recent works include SQLM \cite{chen2025self}, R-Zero \cite{huang2025rzeroselfevolvingreasoningllm}, TT-SI \cite{acikgoz2025selfimprovingllmagentstesttime}, SIRLC \cite{pang2023languagemodelselfimprovementreinforcement}. While this is an efficient approach in data scarce domains where TTT fails, it is not always efficient as there are certain challenging domains which require mastering novel reasoning skills and it is well known that scaling data isn't sufficient in this regimes such as mathematics \cite{hendrycks2021measuringmathematicalproblemsolving}. 

\vspace{0.5cm}
\noindent \textbf{Reinforcement Learning} (RL) is a well established approach for pushing the capabilities of LLMs and recent works such as SEAL \cite{zweiger2025selfadaptinglanguagemodels}, RLAIF \cite{li2025curriculumrlaifcurriculumalignmentreinforcement}, SRLM \cite{yuan2025selfrewardinglanguagemodels} and Memento \cite{zhou2025mementofinetuningllmagents}, which uses a memory-based online RL policy have shown promising potential in the low-cost continual adaptation of LLMs. In RL, \textbf{meta-learning} has been used as well in order train agents in scenarios where it needs to learn novel tasks quickly \cite{gupta2018metareinforcementlearningstructuredexploration}. SOLAR can be seen as thus following meta-learning principles since it learns an adaptation strategy i.e., how to generate effective self weight update using a meta optimization loop. Closely, related are \textbf{self-referential} systems as well which learn to update their own parameters as in \cite{irie2022modernselfreferentialweightmatrix} and \textbf{self-evolving} agents which enable LLM to improvise by autonomously acquiring, refining and learning from experiences generated by the model itself \cite{tao2024surveyselfevolutionlargelanguage, gao2025surveyselfevolvingagentspath}. While RL based approaches are quite good, its often challenging to achieve convergence and design optimal policies which are efficient in terms of compute and time as well.

\vspace{0.5cm}
\noindent 
\textbf{Parameter Generation} is another research direction which has seen several pioneering works such as RPG \cite{wang2025recurrent}, DnD \cite{liang2025drag}, T2L \cite{charakorn2025texttolorainstanttransformeradaption}, ORAL \cite{khan2025oralpromptinglargescaleloras}, COND P-DIFF \cite{jin2024conditionalloraparametergeneration}. DnD generates task-specific parameters from unlabeled prompts without per-task training via a prompt-conditioned hyper-convolutional decoder while T2L does the same but uses a hyper-network and task description instead. ORAL leverages architectural and textual conditioning for flexible, scalable LoRA parameter adaptation. RPG introduces a recurrent diffusion architecture for scalable unconditional LoRA parameter generation. COND P-DIFF applies conditional latent diffusion for controllable LoRA parameter synthesis with strong cross-domain generalization. An associated direction is \textit{model merging} as well, which facilitates generalization to unseen tasks via \textbf{multi-task learning} \cite{shao2025icmfusionincontextmetaoptimizedlora, shao2025incontextmetalorageneration}. While these works have been effective, the limitation is that these are static parameters which once generated do not undergo any further modification but this feature is crucial for domains requiring the implicit meta-knowledge.

\section{Methodology}
\label{methodology}
\begin{figure}[htbp]
    \centering
    \includegraphics[width=\linewidth]{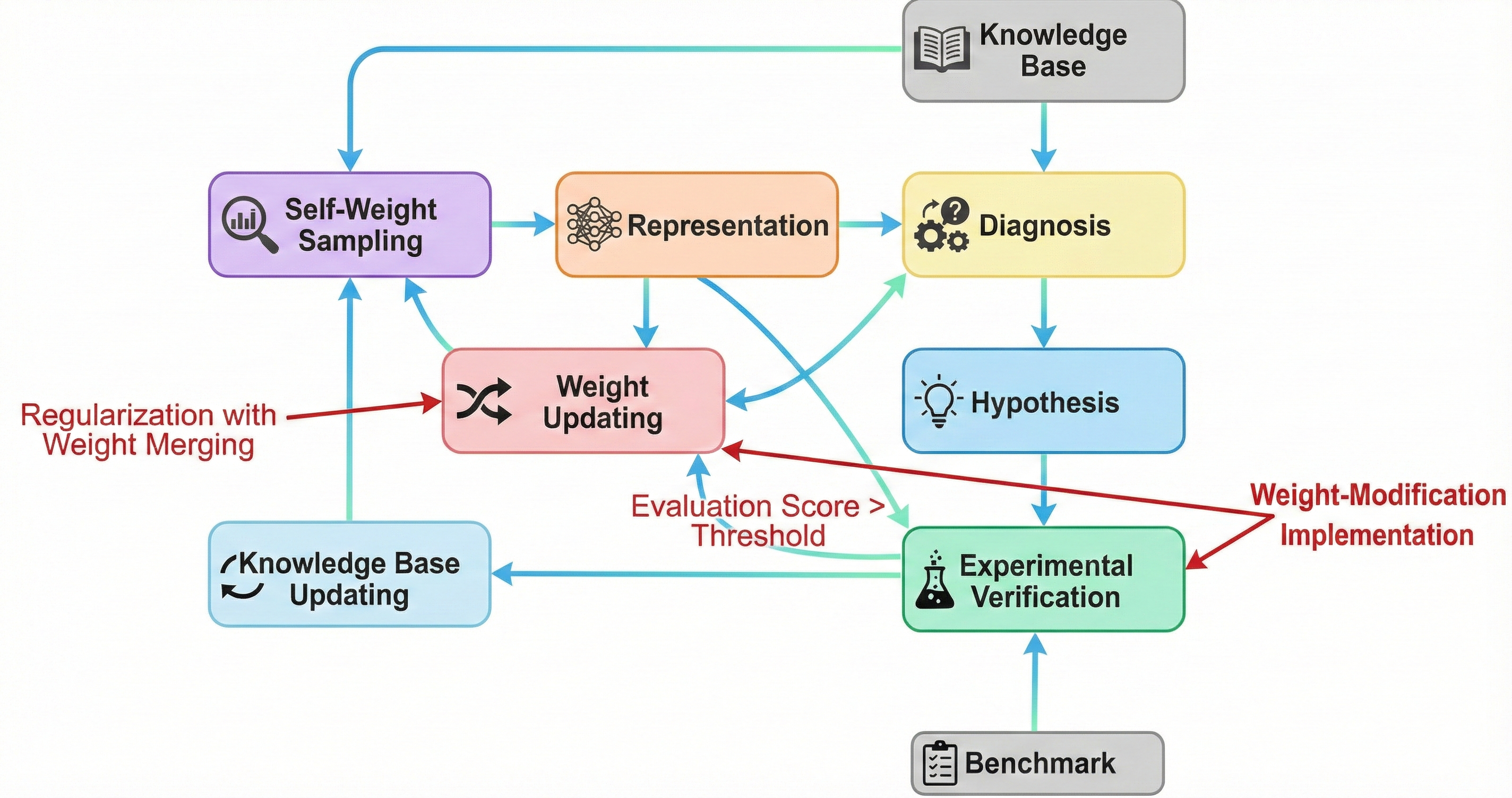}
    \caption{SOLAR's methodology of weight-level meta-knowledge discovery and modification summarized (adapted from \cite{zhang2025toward})}
    \label{fig:GAS}
\end{figure}

In this section, we describe the framework of our proposed approach (see Figure \ref{fig:GAS})\footnote{\cite{zhang2025toward} also provides motivation for the development of self-improving agents at the weight level, however it only provides description of a conceptual framework with no implementation or empirical validation.}. SOLAR starts by treating the LLMs own weights as environment variables to explore, upon which it would systematically propose scientific hypotheses to modify the internal representation space appropriately so as to adapt the LLM to the unseen task. A major challenge for the design, therefore, is the high dimensionality and non-convexity of the LLM weight space itself which makes the initialization and subsequent exploration process extremely complex. To overcome this, we work only with low-rank parameters \cite{hulora} which constitutes a much smaller fraction ($\sim 1\%$) of the original model's weights. In addition, to avoid the limitations arising from selecting a single starting point, which might not be optimal to wiggle around, we prefer to sample from a plausible weight distribution space. This step is essential to eliminate the risk of non-convergence. To get this initial distribution for weights i.e., self-weight sampling, we refer to prior works in large-scale LLM parameter generation and use a convolution-based decoder architecture as the backbone for SOLAR's exploration point initializer.

\vspace{0.5cm}
\noindent Once the weights have been  initialized\footnote{These weights can optionally be encoded into a structured representation correlated with
network performance like world models such as JEPA \cite{lecun2022path}.} for exploration, SOLAR then uses a foundation-model-based agent, which is for now simply an LLM trained using reinforcement learning (RL) to come up with probable hypotheses at inference time for weight-space exploration using test-time scaling and compute. To however, facilitate the training process, it is necessary to first curate by hand a seed knowledge base, consisting of either proven or plausible weight modification strategies, which will then serve as the action space for LLM's initial stages of exploration during RL training. This would be a multi-stage recipe consisting of three distinct progressively harder levels. Level I consists of training the LLM to produce only single valid and efficient self-edits (a self-edit as the name suggests is basically a modification strategy proposed by an LLM to update its own weights depending on the task) from among the ones present in the initial knowledge base. Level II comprises of training the model to output chain-of-self-edits, since coupling strategies sequentially is also helpful (moreover if viewed in a abstract sense, it can be considered in effect as a single complex edit which can be decomposed into simpler instances). Level III is a significantly challenging aspect both for the LLMs as well as from implementation perspective as well, which is basically letting LLMs to explore the hypothesis space in its entirety, thereby going beyond human-crafted approaches. A positive performance in Level III would be a significant leap as it could possibly open up new frontiers in training and fine-tuning paradigms as has been similarly done in other areas as well such as neural architecture search \cite{liu2025alphagomomentmodelarchitecture} and optimization \cite{lu2024discoveringpreferenceoptimizationalgorithms}. 

\vspace{0.5cm}
\noindent 
After plausible hypothesis have been generated by the foundation model-based agent and implemented, its necessary to test the hypothesis. For this purpose, we create a separate evaluation split if available. However, since SOLAR is designed to adapt LLMs efficiently to unseen tasks as well, the dataset for evaluation itself would be generated on the fly using adversarial approaches involving multiple instances of an LLM, one proposing and one solving questions on a particular topic as in SQLM \cite{chen2025self} or R-Zero \cite{huang2025r}. Once the hypothesis has been tested and is found to be valid (as in it improves performance in some pre-determined metric such as accuracy on the \verb|eval| set), it would be then added back into the knowledge base, thereby enriching the action space of LLM for future iterations. In order to prevent catastrophic forgetting, SOLAR implements a meta-level weight regularization technique as well. Therefore, by automating the process of self-improvement using principled methodologies and meta-knowledge in a scientific manner (i.e., propose, validate and accept hypotheses), SOLAR provides a holistic framework towards the next generation of \textbf{AI generating AI} agents, because as soon as web-scale data corpora is exhausted, progress will hinge on a model’s capacity to generate its own high-utility training signal.

\section{Implementation}
\label{sec:implementation}
\subsection{Architecture}

Primary architectural detail in SOLAR's framework is the design of the weight-space exploration initializer. As mentioned in Section \ref{methodology}, we use a convolution based decoder model for this purpose. We assume that we have access to either the unseen task's description or atleast a handful of unlabeled examples representative of its requirements. We then send them to an open-sourced text encoder for embedding extraction. This extraction process can be formally represented as, $c_{i} = \mathrm{Encoder} (p_{i}, \theta),$ where $\mathrm{Encoder} (\cdot, \cdot)$ denotes the embedding extraction function parameterized by $\theta$, and $c_{i}$ represents the extracted embedding corresponding to prompt $p_{i}$. We use an encoder-based language model architecture for this purpose i.e., Sentence-BERT (all-MiniLM-L6-v2 specifically) \cite{reimers2019sentence} \footnote{It is to be noted that BERT's supported sequence length is only 512 and for longer sequences, padding should be done. However, in our use case, maximum sequence length is only 384 and thus padding is not necessary.}. 

\begin{figure}[htpb]
\centering
\includegraphics[width=0.7\linewidth]{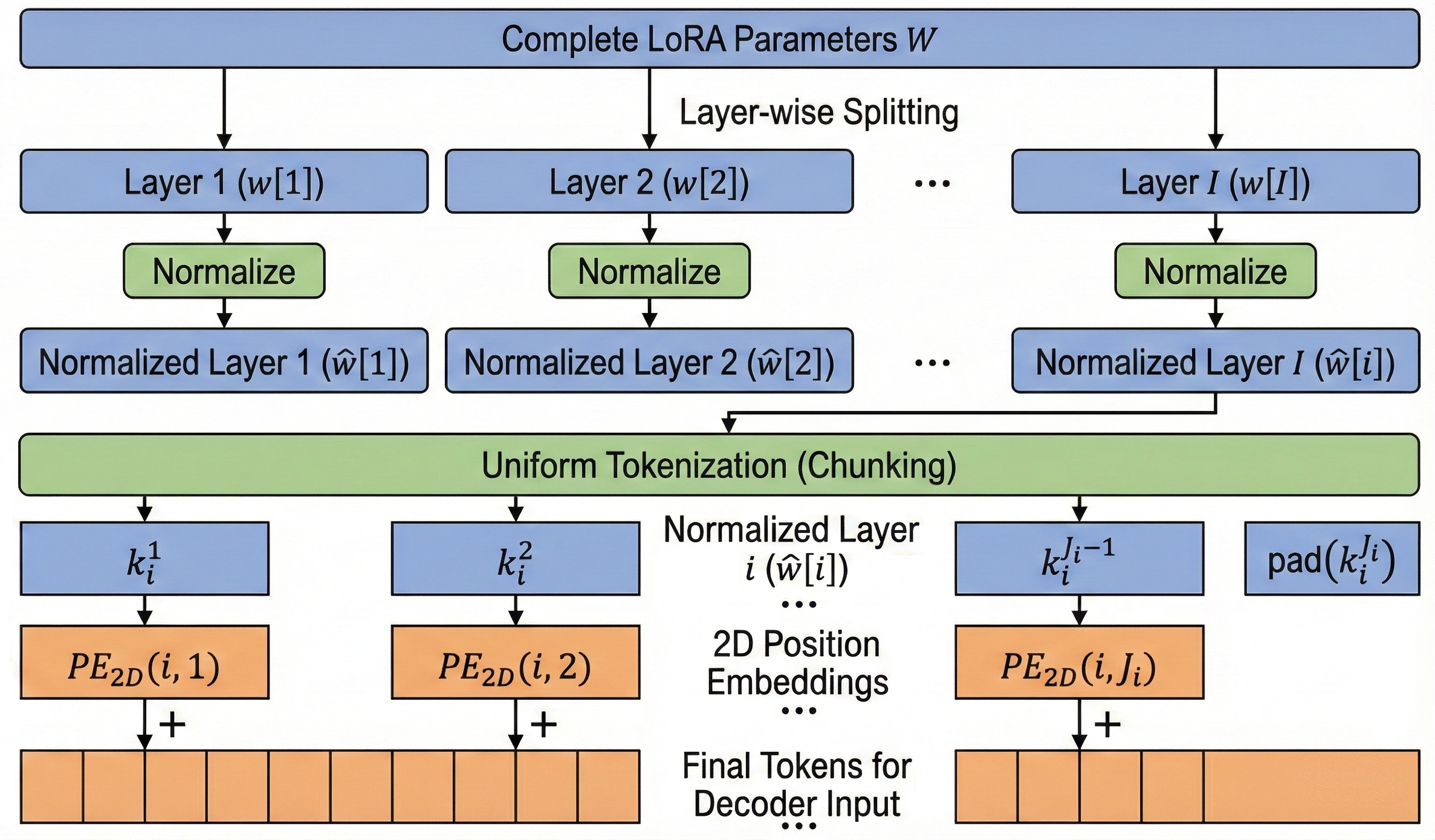}
\caption{Details of the Parameter Tokenization Process}
\label{fig:Parameter_processing}
\end{figure}

\vspace{0.5cm}
\noindent
Next, following \cite{wang2025recurrent}, is the parameter tokenization process (see Figure \ref{fig:Parameter_processing}), which is done so as to preserve both the layer-wise distribution and the cross-layer correlations. Specifically, (i) weights are split according to their layer indices, (ii) layer-wise normalization is applied to mitigate distribution shifts, (iii) parameters are sliced into non-overlapping tokens with uniform size, and (iv) a lightweight permutation state (encoded as a one-hot vector) is used to alleviate symmetry issues \cite{kunin2020neural} when collecting multiple checkpoints. Additionally, 2D position embeddings (first dimension encodes layer index, while second dimension captures the token’s in-layer position). \cite{dosovitskiy2020image} are employed to ensure the network retains positional awareness of each token within the entire set. In our case, each LoRA matrix is of shape $8 \times 896$, which is then split into 7 smaller chunks, each with a shape of $8 \times 128$, which is then finally padded to a uniform size of $10 \times 130$.

\begin{figure}[htbp]
\centering
\includegraphics[width=0.8\linewidth]{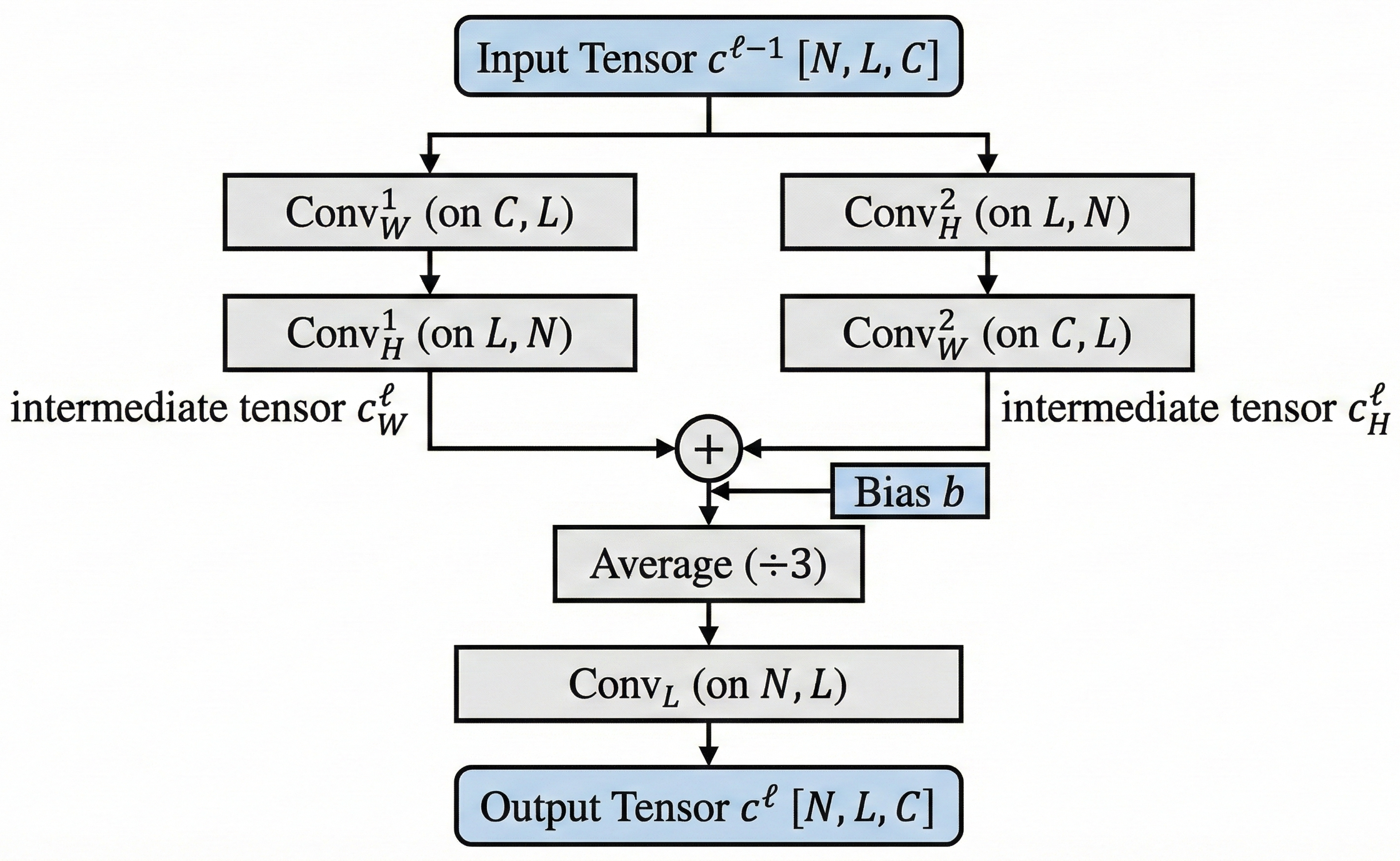}
\caption{Details of the Hyper-Convolutional Decoder Architecture used}
\label{fig:Hyper_convolutional_decoder}
\end{figure}

\vspace{0.5cm}
\noindent
Say, the dimension of prompt embeddings is $[B, N, L, C]$ where $B, N, L \text{ and } C$
denote batch size, length of prompt batch (i.e., number of prompts), sequence length, and hidden dimension, respectively.
The decoder (see Figure \ref{fig:Hyper_convolutional_decoder}) consists of multiple sequential layers, each performing 5 2D convolutions.
These convolutions are divided into three categories: 
i) \textbf{width convolution} that operates on $(C, L)$ dimension, 
ii) \textbf{height convolution} that operates on $(L, N)$ dimension)  
iii) \textbf{layer-wise convolution} that on $(N,L)$ dimension) , with notations $\text{Conv}_{W}$, $\text{Conv}_{H}$, and $\text{Conv}_{L}$.
Each layer consists of two $\text{Conv}_{W}$, two $\text{Conv}_{H}$ and one $\text{Conv}_{L}$.
Given this, the forward operation of the decoder block is,

\begin{equation*}
    \begin{aligned}
        c^{l}_{W} = \text{Conv}^{1}_{H}(\text{Conv}^{1}_{W}(c^{l-1}))
        \\ c^{l}_{H} = \text{Conv}^{2}_{W}\left(\text{Conv}^{2}_{H}(c^{l-1})\right)
        \\
        c^{l} = \text{Conv}_{L}\left(({c^{l}_{W}+c^{l}_{H}+b})/{3}\right)
    \end{aligned}
\end{equation*}
where $c^{l}$ is hidden state output by the $l$ th layer, $c^{0}$ is prompt embedding encoded by the condition extractor, and $b$ is learnable bias.
Through this process, input is transformed from dimension $[B, N, L, C]$ to $[B, N', L', C']$ which is then compatible to be converted into a flattened LoRA adapter for the LLM \footnote{In our present implementation, the entire flow is (128,384,384)  $\to$ (128,200,300) $\to$ (128,100,256) $\to$
(256,50,200) $\to$ (512,50,200) $\to$ (1024,25,200) $\to$ (1024,10,200) $\to$ (2048,10,200) $\to$ (4296,8,128)}. In this work, the base LLM used is Qwen2.5-0.5B-Instruct \cite{qwen2025qwen25technicalreport} and LoRA is applied to the linear projection layers within both the self-attention mechanism and the MLP blocks of the transformer architecture. Specifically, this includes the query, key, value and output projections in attention blocks, as well as the gate, up and down 
projections in MLP blocks.

\subsection{Training}

In this work, we focus on the domain of common-sense reasoning and select 4 datasets for evaluation, namely HellaSwag \cite{zellers2019hellaswagmachinereallyfinish}, BoolQ \cite{clark2019boolqexploringsurprisingdifficulty} as well as the challenge and easy set of AI2 Reasoning Challenge (ARC) \cite{clark2018thinksolvedquestionanswering}. ARC dataset contains grade-school level, multiple-choice science questions. HellaSwag instructs models to select from choices that best finish the sentence among ground truth and an adversarial set of machine-generated wrong answers. BoolQ is a question answering dataset for yes/no questions containing various factual problems. We use existing checkpoints of these datasets \footnote{For training however, even Open-Book Question Answering or OBQA \cite{mihaylov2018can}, Physical Interaction: Question Answering or PIQA \cite{bisk2019piqareasoningphysicalcommonsense} and WinoGrande \cite{sakaguchi2019winograndeadversarialwinogradschema} have been used as well. OBQA aims to promote research in advanced question-answering with salient facts summarized as an open book. PIQA focuses on everyday situations with a preference for a typical solutions. WinoGrande features a fill-in-a-blank task with binary
options for commonsense reasoning questions.} (batch size was 32 and number of samples was 5000) which have been collected by first pretraining on the target dataset for 75 steps with a learning rate of 1e-4 and then performing fine-tuning on the target dataset for 50 additional steps with a learning rate of 1e-5, while saving a checkpoint at each step.

\vspace{0.5cm}
\noindent 
Subsequently, prompt-checkpoint pairing is done as follows. Given a dataset $P$, it is first divided it into non-overlapping prompt batches 
$[p_{1}, \cdots, p_{i}, \cdots, p_{I}]$. Denote the trained LLM checkpoints of this dataset as $M = [m_{1}, \cdots, m_{j}, \cdots, m_{J}]$. Then randomly  a batch of prompts and a corresponding checkpoint is picked to create a pair $\{p_{i}, m_{j}\}$, which then serves as an input-output data point for training the decoder. The objective function for training is the mean squared error (MSE) loss between the output from the decoder's last block for a particular prompt batch and the training checkpoint associated with it. 

\vspace{0.5cm}
\noindent Next crucial step is the hand-crafting of seed knowledge base. To this end, we identify five primary families of strategies\footnote{Unfortunately, there are no research works highlighting approaches for optimizing the performance of LoRA's obtained via the process of parameter generation, thereby posing a major challenge in identification of plausible strategies, which had to be cherry-picked via trial and error.}, each containing its own sub-strategies as well, namely

\begin{itemize}
    \item Test-Time Training (TTT) using  input perplexity minimization \textbf{\cite{hu2025testtimelearninglargelanguage}} or via reinforcement learning \cite{zuo2025ttrltesttimereinforcementlearning} for example by using self-reflection and verification loops like GEPA \cite{agrawal2025gepareflectivepromptevolution}, ReflectEvo \cite{li2025reflectevoimprovingmetaintrospection}, REVISE \cite{lee2025reviselearningrefinetesttime} or Instruct-of-Reflection \cite{liu2025instructofreflectionenhancinglargelanguage}. It could also involve prompt optimization using frameworks like TextGrad \cite{yuksekgonul2024textgradautomaticdifferentiationtext} or CAST \cite{tang2025enhancingcrosstasktransferlarge}
    \item Post-training data-free LoRA modifications such as mixing LoRA subspaces obtained by weight decomposition of constituent matrices \textbf{\cite{wu2025mixtureofsubspaceslowrankadaptation}} or bounding norm of selected parameters \cite{wang2025normboundedlowrankadaptation} or evening merging multiple task-specific LoRA adapters \cite{zhao2024mergingloraslikeplaying}
    \item SQLM \cite{chen2025selfquestioninglanguagemodels}, R-Zero \textbf{\cite{huang2025rzeroselfevolvingreasoningllm}} or SEAL \cite{zweiger2025selfadaptinglanguagemodels} like reinforcement learning based frameworks which enable LLMs to self-adapt by generating their own finetuning data and update directives (another example is TT-SI \cite{acikgoz2025selfimprovingllmagentstesttime})
    \item Test-Time Scaling (TTS) using either a router or an ensemble approach i.e., we generate and perform inference with multiple adapters obtained by using different representative prompt batches and to obtain the final prediction, select either the most confident
    prediction (max\_confidence) or by a majority vote or sum\_logprobs (i.e., sum log probabilities across
   adapters per prediction and pick the one with highest total logprob)
   \item Latent Space (LS) Approaches which aim at working or modifying the internal layers \cite{hu2025slotsamplespecificlanguagemodel} or hidden activations \cite{zhang2025latentevolveselfevolvingtesttimescaling} directly of the LLM. It may also involve decoding algorithms which modify the sampling procedure itself \cite{karan2025reasoningsamplingbasemodel,wang2025endmanualdecodingtruly}. We consider them as part of latent space family because they tamper with internal probability distribution of next-tokens unlike other families which modify the parameters explicitly. 
\end{itemize}

\noindent We first formulate the objective for outer-loop RL training which generates adaptation strategies $\texttt{AS}$, as in \cite{zweiger2025selfadaptinglanguagemodels}.
Let $\theta$ denote the parameters of the language model $\texttt{LM}_\theta$. In order to adapt to an unseen dataset (task) $\mathcal{D}$, SOLAR requires as specified in Section \ref{methodology}, $C$ which is a context containing information relevant to the task and $\tau$ which is the evaluation strategy and metric used to assess the model's downstream adaptation. Based on $C$, SOLAR generates an $\texttt{AS}$ and updates its parameters accordingly $\theta' \leftarrow \texttt{Update}(\theta, \texttt{AS})$. We thus have an RL setup i.e., the model takes an \textit{action} (generating $\texttt{AS}$), receives a \textit{reward} $r$ based on $\texttt{LM}_{\theta'}$'s performance on $\tau$ and updates its policy to maximize expected reward,
\vspace{-1pt}
\begin{equation*}
\label{eqn:objective}
\mathcal{L}_{\text{RL}}(\theta_t) :=\, 
 -\mathbb{E}_{(C, \tau) \sim \mathcal{D}} \left[ 
    \mathbb{E}_{\texttt{AS} \sim \text{LM}_{\theta_t}(\cdot \mid C)} 
    \left[ r(\texttt{AS}, \tau, \theta_t) \right] 
\right]
\end{equation*}

\noindent It is to be noted that the reward assigned to a given action depends on the model parameters $\theta$ at the time the action is taken (since $\theta$ is updated to $\theta'$, which is then evaluated). An implication of this is that the while modeling the RL state, one must therefore include $\theta$ in the policy's parameters as well along with $C$, even though the policy's observation is limited to $C$ (because it is extremely infeasible to directly place $\theta$ in the LLM's context window). Therefore, the (state, action, reward) triples which have been collected by using an older model weights, $\theta_{\text{old}}$, will not be aligned for the current model $\theta_{\text{current}}$. Hence, an on-policy approach should be adapted, by which adaptation strategies are sampled from and, even more importantly, the rewards itself will be calculated using the current model.

\vspace{0.5cm}
\noindent In particular, the specific on-policy approach used is ReST$^\text{\textit{EM}}$ \cite{singh2024humandatascalingselftraining} where samples are first generated\footnote{Currently, only a deterministic number of samples are being generated, 15 to be precise. This could however be improvised to be dynamic in future version of the work wherein samples would continue to be generated until a particular confidence threshold, as determined by the model itself is reached instead. The same is true for number of iterations as well which is just 2 for now.} from the current model and are filtered by using binary feedback [$r(\texttt{AS}, \tau, \theta_t)$ is 1 if on $\tau$, $\texttt{AS}$ improves $\text{LM}_{\theta_t}\text{'s performance}$ and is 0 otherwise]. The model is then fine-tuned on these samples and this continues in an iterative manner (See Algorithm \ref{alg:SOLAR_sequential}).

\vspace{0.5cm}
\noindent
A subtle detail, which hasn't yet been covered is the exact nature of the adaptation strategy itself. This depends on the particular strategy family being used, however the format is consistent across all which is basically a JSON object specifying the particular configurations to be used\footnote{Since the model being used is Qwen2.5-0.5B-Instruct, it was facing difficulty in following instructions given in the prompt for generation of structured outputs even after temperature alteration. In such cases, verification and formatting was done by using Qwen2.5-7B-Instruct instead.}. It contains a field, \verb|family| which takes values \verb|TTT|, \verb|LoRA| and \verb|TTS|. Currently, the following choices have been experimented

\begin{figure}[htpb]
    \centering
    \includegraphics[width= \linewidth]{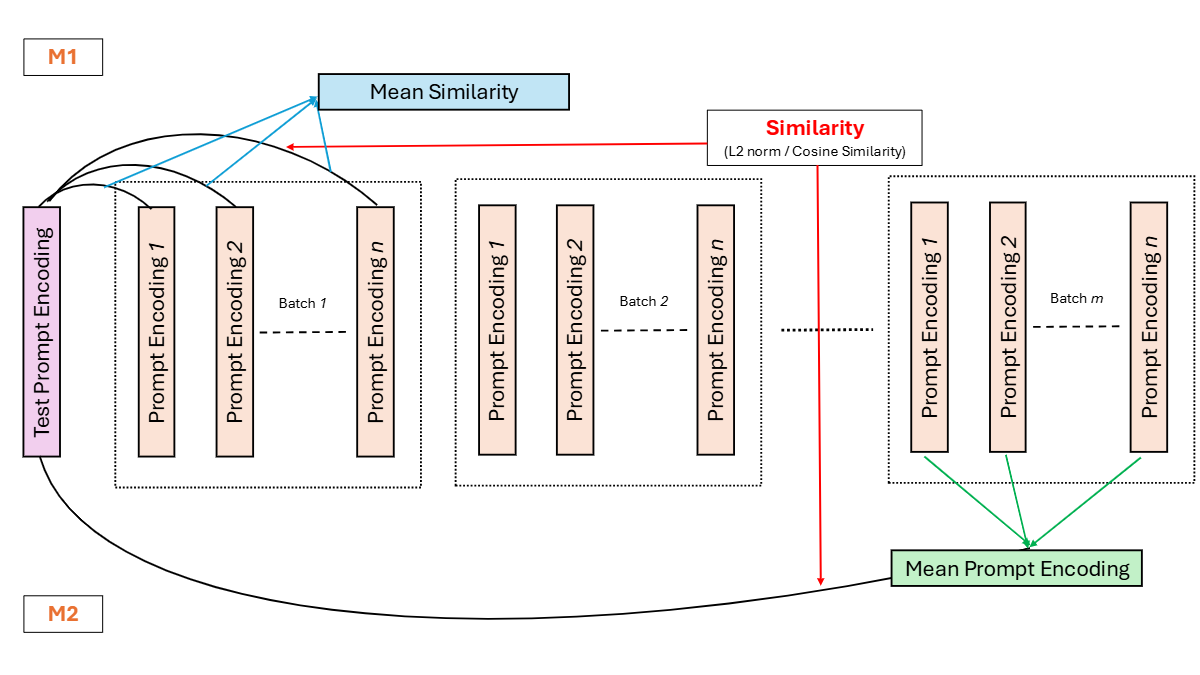}
    \caption{Router Approach for TTS }
    \label{fig:Router}
\end{figure}

\noindent 
 
\begin{itemize}
    \item For \verb|TTT|, we use \cite{hu2025testtimelearninglargelanguage} and the corresponding JSON object has fields  
    \verb|ttl_steps| (number of training steps in the TTL loop), 
    \verb|learning_rate|,
    \verb|batch_size| and
    \verb|shuffle_data| (boolean variable). 
    \item For \verb|LoRA| modifications, we use two-subspace (TS) mixing version from \cite{wu2025mixtureofsubspaceslowrankadaptation} and the corresponding JSON object has only a single field, namely \verb|lambda| which is a hyperparameter determining the ratio in which the two resulting subspaces must be mixed.
    \item For \verb|TTS|, we use either an ensemble or router approach. In the router approach (see Figure \ref{fig:Router}), we basically sample multiple prompt batches and choose that batch whose average of  similarity scores\footnote{Cosine similarity and Euclidean distance were tested and the latter was found to perform better empirically. Thus, $\texttt{avg\_sim\_score}$ and $\texttt{avg\_prompt\_embed}.$ use euclidean distance by default. Alternatively, measure of similarity can also be included as a new field but hasn't been explored in the current work.} of individual prompts (\textcolor{orange}{M1}) or averaged prompt embedding (\textcolor{orange}{M2}), is closest to that of the question at test time. The corresponding JSON object has fields \verb|num_prompt_batches| (indicating the number of prompt batches to be sampled from the test split of unseen dataset) and \verb|method| which can take one of five values - \verb|avg_sim_score|, \verb|avg_prompt_embed|,  \verb|max_confidence|, \verb|majority_vote| or (summing log probabilities) i.e., \verb|sum_logprobs| (former two belong to router approach and the latter three constitute the ensemble approach).
    \item For \verb|LS|, we use \cite{hu2025slotsamplespecificlanguagemodel} and the corresponding JSON object has fields  
    \verb|times| and \verb|learning_rate|.
\end{itemize}

\begin{algorithm}[!t]
\small
\caption{Sequential Multi-Level RL Loop for Adaptation Strategy (AS) Generation of SOLAR}
\label{alg:SOLAR_sequential}
\begin{algorithmic}[1]
\STATE \textbf{Input:} Base LM$_\theta$, dataset context $C$, evaluation metric $\tau$, initial knowledge base $K$
\STATE \textbf{Init:} Low-rank adapter generator $G$, sampled adapters $S \leftarrow \texttt{Sampler}(C, G)$
\medskip
\STATE \textbf{Level I (Single-edit self-training)}:
\FOR{iteration $t=1,\dots,T_1$}
    \STATE Propose single-edit AS from $K$ \hfill{$\texttt{AS} \sim \text{LM}_\theta(K,C)$}
    \STATE Apply AS and obtain weights \hfill{$\theta' \leftarrow \texttt{ApplyStrategy}(\theta,\texttt{AS},S)$}
    \STATE Evaluate \hfill{$\texttt{Ans}\sim\text{LM}_{\theta'}(\cdot\mid\tau)$}
    \STATE Compute reward \hfill{$r\leftarrow r(\texttt{Ans},\tau)$}
    \IF{$r > \text{threshold}_1$}
        \STATE $\theta \leftarrow \texttt{RL\_Update}(\theta,r,\texttt{AS})$
    \ENDIF
\ENDFOR
\medskip
\STATE \textbf{Level II (Chained/compositional strategies)}:
\FOR{iteration $t=1,\dots,T_2$}
    \STATE Propose chain of edits \hfill{$\texttt{AS}=[e_1,\dots,e_k],\; e_i\in K$}
    \STATE Sequentially apply chain \hfill{$\theta_0\leftarrow\theta$; \; $\theta_i\leftarrow\texttt{ApplyStrategy}(\theta_{i-1},e_i,S)$}
    \STATE Evaluate final weights \hfill{$\texttt{Ans}\sim\text{LM}_{\theta_k}(\cdot\mid\tau)$}
    \STATE Compute reward \hfill{$r\leftarrow r(\texttt{Ans},\tau)$}
    \IF{$r > \text{threshold}_2$}
        \STATE Add chain to KB \hfill{$K \leftarrow K \cup \{\texttt{AS}\}$}
        \STATE $\theta \leftarrow \texttt{RL\_Update}(\theta,r,\texttt{AS})$
    \ENDIF
\ENDFOR
\medskip
\STATE \textbf{Level III (Open-ended exploration)}:
\FOR{iteration $t=1,\dots,T_3$}
    \STATE Generate unconstrained AS \hfill{$\texttt{AS}\sim\text{LM}_\theta(\cdot\mid C)$ (novel structure)}
    \STATE Validate (syntax/safety); if invalid \textbf{continue}
    \STATE Apply AS conservatively (strong meta-reg) \hfill{$\theta' \leftarrow \texttt{ApplyStrategy}(\theta,\texttt{AS},S)$}
    \STATE Evaluate and compute reward \hfill{$\texttt{Ans}\sim\text{LM}_{\theta'}(\cdot\mid\tau)$; \; $r\leftarrow r(\texttt{Ans},\tau)$}
    \IF{$r > \text{threshold}_3$}
        \STATE $K \leftarrow K \cup \{\texttt{AS}\}$; \; $\theta \leftarrow \texttt{RL\_Update}(\theta,r,\texttt{AS})$
    \ELSE
        \STATE Penalize harmful proposals in policy update
    \ENDIF
\ENDFOR
\STATE \textbf{Return:} Refined parameters $\theta^*$, enriched KB $K^*$
\end{algorithmic}
\end{algorithm}

\section{Experiments}
\label{sec:experiments}

\subsection{Setup}
As described in Section \ref{sec:implementation}, the base LLM used is Qwen2.5-0.5B-Instruct, domain is common-sense-reasoning and evaluation datasets are ARC-c, BoolQ, HellaSwag and ARC-e. Baselines used include quite recent works such as DnD \cite{liang2025drag}, Test-Time Learning (TTL) \cite{hu2025testtimelearninglargelanguage}, Decoupled and Orthogonal Merging (DOM)\footnote{DOM is a data-free framework for LoRA merging. It separates parameters into magnitude and direction components and merges them independently, thereby reducing the impact of magnitude differences on the directional alignment of the merged models, thus helping in preserving task information. It also uses a data-free, layer-wise gradient descent method with orthogonal constraints to mitigate interference during the merging of direction components. For evaluation on a target dataset, LoRA's of remaining datasets are merged and used.} \cite{zheng2025decoupleorthogonalizedatafreeframework} and average of task-specific training LoRA's \cite{hulora}. On one extreme, TTL uses the entire unlabeled corpus of the training LoRA's in addition to the 128 unlabeled examples from the target dataset as seen by SOLAR. On the other extreme, instead of using the unlabeled corpus, DOM merges all the 7 training LoRA's inclusive of the target set. 

\subsection{Hardware}
All experiments were conducted on a high-performance computing node running Ubuntu 22.04.1. The backend processor was EPYC 8434P, which had 48 physical cores (96 logical threads), 256 GB of system RAM and a maximum clock speed of 2.5 GHz. Four NVIDIA RTX A6000 GPUs, each with 48 GB of dedicated VRAM were utilized. Python version used was 3.12.11 and GPU-accelerated tasks were managed using CUDA version 12.4.

\subsection{Results}

The major results of this work are presented in Table \ref{tab:Main_Results_DoubleModel} wherein we conduct experiments of 5 benchmarks which are in the domain of common-sense reasoning and also on 5 out-of-domain benchmarks namely GSM-MC and MATH-MC \footnote{GSM-MC and MATH-MC are multiple choice versions of the standard GSM-8K \cite{cobbe2021training} and MATH \cite{hendrycks2021measuring} datasets. They were selected for two reasons - ease of evaluation and correlation with performance on their subjective counterparts \cite{zhang2024multiple}.} to evaluate mathematical reasoning, DivLogicEval \cite{chung2025divlogiceval} for logical reasoning, SocialIQA \cite{sap2019socialiqa} for reasoning about social interactions and CodeMMLU \cite{manh2024codemmlu} for reasoning about code-related tasks. It can be seen that SOLAR in its initial version itself outperforms the task-specific training LoRA's, TTL, DOM and even DnD by a significant margin, showcasing the promising potential it is capable of, if further levels of RL training\footnote{This might be quite time-intensives however with current version itself taking around 4 days using 2 A6000 GPU's. The reason for using only 2 despite 4, is because Qwen family has 14 attention heads and the vllm serves used for improved efficiency in inference requires this number to be divisible by the number of GPU's which is only possible if either 2 or 7 are available.} is completed as well.

\begin{table*}[!t]
\centering
\setlength{\tabcolsep}{4pt}
\begin{tabular}{c|ccccc|c|ccccc|c}
\hline
\hline
\multicolumn{1}{c|}{} & \multicolumn{5}{c|}{\textbf{In-Domain Tasks}} & \textbf{Avg.} & \multicolumn{5}{c|}{\textbf{Out-of-Domain Tasks}} & \textbf{Avg.} \\
\cline{2-13}
\textbf{Dataset} & ARC-e & ARC-c & BoolQ & HellaSwag & PIQA & \textbf{In} & GSM & MATH & Logic & Social & Code & \textbf{Out} \\
\hline
\rowcolor{lightred}
LoRA & 47.4 & 39.7 & 14.7 & 26.3 & 51.5 & 35.9 & 15.6 & 6.8 & 20.3 & 39.5 & 29.8 & 22.4 \\
\rowcolor{lightgray}
TTL & 24.4 & 24.7 & 44.4 & 25.9 & 51.9 & 34.3 & 23.5 & 19.7 & 26.2 & 34.9 & 29.7 & 26.8 \\
\rowcolor{lightgray}
DOM & 56.5 & 38.9 & 33.2 & 28.3 & 18.8 & 35.1 & 17.7 & 2.6 & 24.7 & 51.3 & 31.6 & 25.6 \\
\rowcolor{lightgray}
DnD & 70.9 & 48.1 & 51.9 & 26.5 & 47.8 & 49.0 & 20.8 & 24.1 & 21.0 & 33.5 & 29.1 & 25.7 \\
\rowcolor{lightgreen}
\textbf{SOLAR} & \textbf{74.7} & \textbf{55.5} & \textbf{58.8} & \textbf{48.3} & \textbf{60.1} & \textbf{59.5} & \textbf{30.3} & \textbf{24.5} & \textbf{25.1} & \textbf{55.0} & \textbf{35.6} & \textbf{34.1} \\
\hline
$\Delta$ DnD & 3.8\greenup & 7.4\greenup & 6.9\greenup & 21.8\greenup & 12.3\greenup & 10.4\greenup & 9.5\greenup & 0.4\greenup & 4.1\greenup & 21.5\greenup & 6.5\greenup & 8.4\greenup \\
$\Delta$ DOM & 18.2\greenup & 16.6\greenup & 25.6\greenup & 20.0\greenup & 41.3\greenup & 24.3\greenup & 12.6\greenup & 21.9\greenup & 0.4\greenup & 3.7\greenup & 4.0\greenup & 8.5\greenup \\
$\Delta$ TTL & 50.3\greenup & 30.8\greenup & 14.4\greenup & 22.4\greenup & 8.2\greenup & 25.2\greenup & 6.8\greenup & 4.8\greenup & 1.1\reddown & 20.1\greenup & 5.9\greenup & 7.3\greenup \\
$\Delta$ LoRA & 27.3\greenup & 15.8\greenup & 44.1\greenup & 22.0\greenup & 8.6\greenup & 23.6\greenup & 14.7\greenup & 17.7\greenup & 4.8\greenup & 15.5\greenup & 5.8\greenup & 11.7\greenup \\
\hline
\hline
\end{tabular}
\caption{Accuracy (in \%) of SOLAR Level I approach over the baselines TTL (25.2\greenup), LoRA (23.6\greenup), DOM (24.3\greenup), and DnD (10.4\greenup) for in-domain tasks, and TTL (7.3\greenup), LoRA (11.7\greenup), DOM (8.5\greenup), and DnD (8.4\greenup) for out-of-domain tasks, where values in parentheses denote average $\Delta$ (change in accuracy) for \textbf{Qwen2.5-0.5B-Instruct}.}
\label{tab:Main_Results_DoubleModel}
\end{table*}

\noindent Following were the adaptation strategies identified, which enabled SOLAR to reach the accuracy levels presented, 
\begin{itemize}
    \item For ARC-e and PIQA, it was \verb|TTT| family with configuration \{``\verb|ttl_steps|'': 25,   "\verb|learning_rate|'': 1e-5,
    "\verb|batch_size|'': 4, 
    "\verb|shuffle_data|'': True\}
    \item For ARC-c and SocialIQA, it was \verb|LS| family with configuration \{``\verb|times|'': 5,   "\verb|learning_rate|'': 0.1\}
    \item For BoolQ, GSM-MC and MATH-MC, it was \verb|LoRA| family with TS-mixing strategy and the configuration was \{``\verb|lambda|'': \verb|0.5|\}
    \item For HellaSwag, DivLogicEval and CodeMMLU, it was \verb|TTS| family. Ex:, for Hellaswag, the corresponding configuration was \{``\verb|num_prompt_batches|'': \verb|20|, "\verb|method|'': \verb|max_confidence|\}, indicative of the ensemble approach
\end{itemize}

\subsection{Ablation Study}

\begin{figure}
    \centering
    \includegraphics[width=\linewidth]{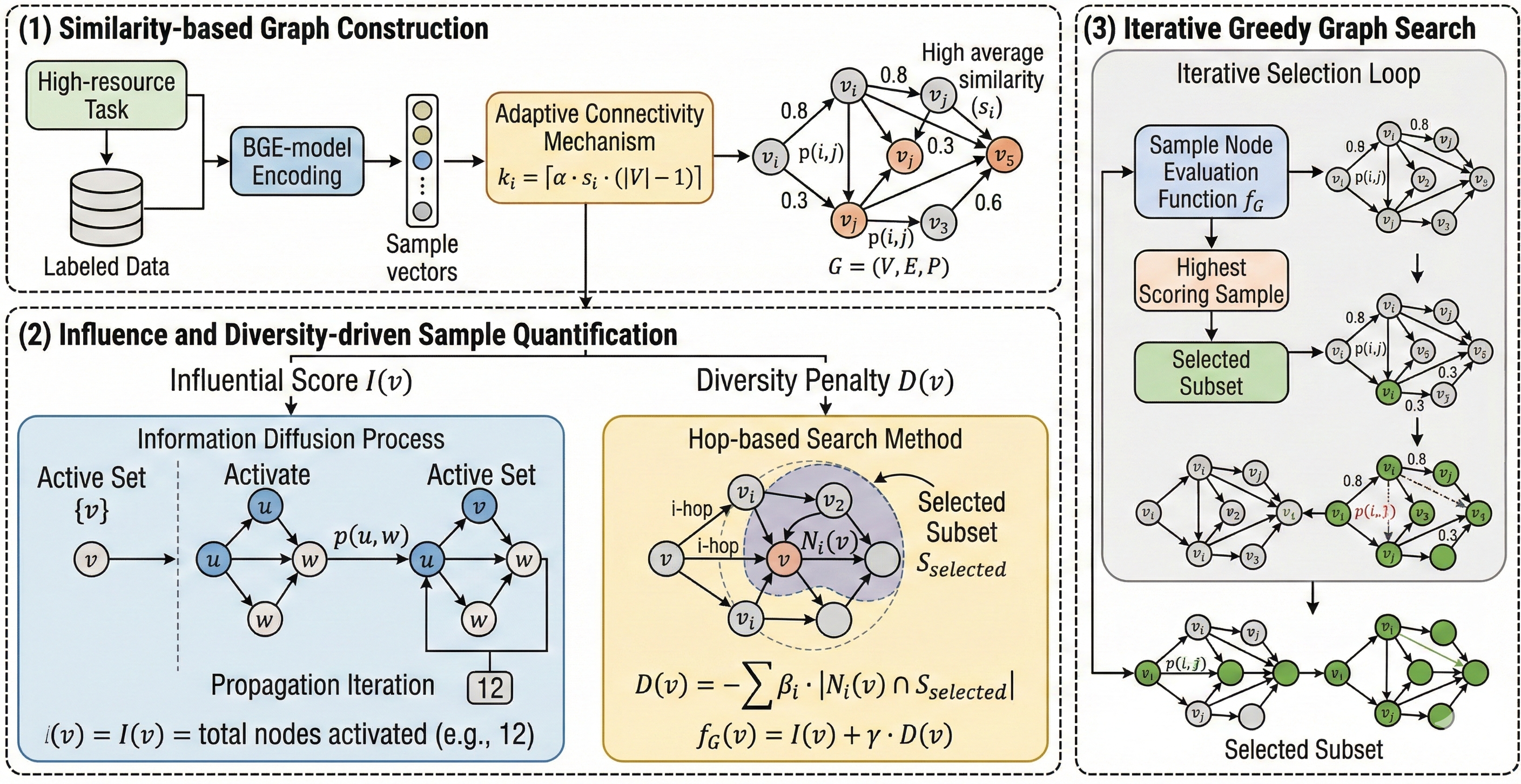}
    \caption{Details of the Prompt Selection Strategy used in Ablation Study}
    \label{fig:CAST}
\end{figure}

A primary effect we would like to isolate and study is that of initial prompt batch provided to start the LLM adaptation process using SOLAR. It would be ideal if SOLAR results in similar performance even if a highly representative, diverse and influential prompt batch is used. For this purpose, inspired by \cite{tang2025enhancingcrosstasktransferlarge}, we use the following strategy for prompt filtering and selection (see Figure \ref{fig:CAST}). 

\vspace{0.5cm}
\noindent We first model inter-prompt relations as a directed graph \(\mathcal G = (\mathbf V,\mathbf E,\mathbf P)\), wherein each prompt is encoded as a vector by using Sentence-BERT.  Each vertex \(v_i\in \mathbf V\) denotes a prompt (sample), a directed edge \(e(i,j)\in \mathbf E\) connects \(v_i\) to its neighbor \(v_j\),  
and weight \(p(i,j)\in\mathbf P\) is the cosine similarity of their embeddings.  For each node \(v_i\), an $s_i$ is computed as shown below so that nodes with higher average similarity make more connections.
\[
s_i = \frac{1}{|\mathbf V|-1}\sum_{j\neq i} s(i,j), \quad
k_i = \left\lceil \alpha \cdot s_i \cdot (|\mathbf V| - 1) \right\rceil
\]

\vspace{0.5cm}
\noindent
Samples are then scored by by (1) influence and (2) diversity. The influence score \(I(v)\) is obtained by a diffusion simulation\footnote{The simulation is run 20 times and is then averaged to obtain the final value.}. For this, first initialize an active set \(S_{\rm active}=\{v\}\), then iteratively sample an active node \(u\) and attempt to activate each neighbor \(w\in N_1(u)\) with probability \(p(u,w)\).  Newly activated nodes join \(S_{\rm active}\).  This process is repeated until no active nodes remain. Let \(I(v)\) be the total number of visited nodes. Diversity penalty \(D(v)\) measures overlap with already selected nodes:
\[
D(v) = -\sum_{i=1}^k \beta^i \,\bigl|N_i(v) \cap S_{\rm selected}\bigr|,
\quad
f_{\mathcal G}(v) = I(v) + \gamma\,D(v)
\]

\vspace{0.5cm}
\noindent Finally, greedy graph search is done to select the final prompt subset $S$. For this, start with \(S=\emptyset\) and at each round pick
\[
v^* = \arg\max_{v \notin S} f_{\mathcal G}(v),
\] 
\(v^*\) is then added to \(S\) and diversity penalties only for neighbors of \(v^*\) are updated\footnote{Note that the influence scores are precomputed.}. This process continues until \(|S|\) reaches the target size which in our case is 128.

\vspace{0.5cm}
\noindent Fortunately, the influence of the initial prompt batch was \textbf{marginal} (\textit{with just a \textbf{0.3}\% improvement in accuracy when averaged across all evaluation datasets}), indicating that SOLAR can efficiently adapt LLMs to unseen datasets without the requirement of high-quality or manually curated dataset. Only a handful of unlabeled prompt instances which are merely indicative of the task suffice.

\section{Conclusion}
In this paper, we introduce SOLAR which is a novel paradigm for Streaming and Continual Learning by empowering LLMs to autonomously discover and retain parameter-
level adaptation strategies. By bridging the gap between rapid test-time adaptation (plasticity) and long-term meta-knowledge retention (stability), SOLAR addresses the core challenges of deploying agents in non-stationary environments. While currently reliant on a seed knowledge base,
the framework lays the groundwork for fully autonomous, self-evolving systems capable of navigating the open-ended
drifts of the real world. Another key tradeoff is that of real-
time adaptation versus computation. While SOLAR's training phase is compute-intensive, the inference-time application of learned strategies is rapid. By pre-compiling complex adaptation routines into the knowledge base, SOLAR shifts
the computational burden from the streaming phase to the offline meta-learning phase. This allows the agent to react to concept drift in near real-time by simply retrieving and applying a cached strategy, rather than performing expensive gradient descent from scratch every time.

\section{Acknowledgments}
The authors would like to thank Professor Sashikumaar Ganesan, from the Department of Computational and Data Science at Indian Institute of Science, Bangalore for feedback and additional compute resources required to execute this project.

\section*{Declaration on Generative AI}

During the preparation of this work, the authors used Large Language Models (GPT-5.2, Claude Opus 4.5 and Gemini-3) as a writing assistant tool for drafting content, to generate literature review, for abstract drafting, to paraphrase and reword, to improve writing style, for grammar and spelling check as well as to generate the images used in the paper. The process was interactive. After writing the core content, the authors used LLMs with specific prompts
to refine the text. These prompts included requests to “check for grammatical errors,” “rephrase
this sentence for clarity,” “make this paragraph more concise,” or “suggest alternative phrasing to
improve flow.” The LLMs were not used to generate any scientific ideas, experimental results, data analysis or
other core intellectual contributions of the paper. After using these tool(s)/service(s), the authors reviewed and edited the content as needed and take full responsibility for the publication’s content.

\bibliography{aaai2026}

\end{document}